\documentclass{article}

\usepackage{microtype}
\usepackage{graphicx}
\usepackage{subfigure}
\usepackage{booktabs} % for professional tables
\usepackage{pgfplots}

\usepackage{hyperref}
\pgfplotsset{compat=1.18}

\usepackage[accepted]{icml2025}

\usepackage{amsmath}
\usepackage{amssymb}
\usepackage{mathtools}
\usepackage{amsthm}
\usepackage{float}      % For [H] table/figure positioning
\usepackage{tikz}       % For creating graphics (required by pgfplots)
\usepackage{pgfplots}   % For generating the bar chart

\usepackage[capitalize,noabbrev]{cleveref}

\theoremstyle{plain}

\theoremstyle{definition}

\theoremstyle{remark}

\usepackage[textsize=tiny]{todonotes}

\begin{document}

\twocolumn[
\icmltitle{A Comprehensive Evaluation of Transformer-Based Question Answering Models and RAG-Enhanced Design}



\icmlsetsymbol{equal}{*}

\begin{icmlauthorlist}
\icmlauthor{\textbf{Zichen Zhang}}
\icmlauthor{\textbf{Hongwei Ruan}} \hspace*{3em}
\icmlauthor{Kunlong Zhang}
\icmlauthor{Yiming Luo}
\end{icmlauthorlist}

% \icmlaffiliation{sch}{University of Michigan, Ann Arbor}

\icmlcorrespondingauthor{Firstname1 Lastname1}{first1.last1@xxx.edu}
\icmlcorrespondingauthor{Firstname2 Lastname2}{first2.last2@www.uk}

\icmlkeywords{Machine Learning, ICML}

\vskip 0.3in
]




% \begin{abstract}
% This document provides a basic paper template and submission guidelines.
% Abstracts must be a single paragraph, ideally between 4--6 sentences long.
% Gross violations will trigger corrections at the camera-ready phase.
% \end{abstract}

\section{Introduction}
\label{submission}

Question Answering (QA) is an essential and rapidly advancing area within Natural Language Processing (NLP), focusing on models capable of extracting or generating precise answers from textual information. The advent of Transformer-based architectures, exemplified by models such as BERT and the Qwen series, has dramatically improved QA accuracy and efficiency \cite{BERT,qwen}. These advanced models typically undergo pre-training on extensive text corpora and fine-tuning using widely recognized datasets, including the Stanford Question Answering Dataset (SQuAD) \cite{SQuADQA}. Nonetheless, despite significant progress, reliably retrieving contextually relevant information, especially in multi-hop retrieval tasks, remains challenging, motivating the development of Retrieval-Augmented Generation (RAG) systems \cite{Facebook}.

\subsection{Task Description}
This project focuses explicitly on multi-hop QA tasks, employing Retrieval-Augmented Generation (RAG) methodologies to enhance the accuracy and robustness of QA systems. The core objective is to systematically benchmark established retrieval methods, such as cosine similarity and Maximal Marginal Relevance (MMR) \cite{MMR}, and introduce advanced hybrid retrieval techniques \cite{sawarkar2024blended}. Our hybrid retrieval method uniquely integrates dense vector similarity with keyword-based lexical matching, balancing document relevance and diversity. Furthermore, we explore query optimization techniques aimed at improving retrieval effectiveness by breaking complex queries into more manageable sub-queries \cite{zhuang2024efficientragefficientretrievermultihop}, thereby enhancing multi-hop reasoning capabilities.

\subsection{Problem Significance}
Multi-hop QA has emerged as a critical capability required by contemporary intelligent systems, particularly digital assistants, automated support platforms, and sophisticated educational tools. The significance of enhancing multi-hop retrieval lies in its direct impact on system usability, user satisfaction, and the overall efficacy of knowledge extraction in complex, real-world scenarios. By addressing the inherent limitations of conventional retrieval approaches and optimizing retrieval strategies, such as Maximal Marginal Relevance \cite{MMR}, hybrid retrieval \cite{sawarkar2024blended}, and query optimization pipeline, our project seeks to substantially improve QA performance. This advancement can provide significant benefits in practical applications, delivering precise and contextually accurate responses to intricate user queries, thus improving user experience and information reliability.

\subsection{Challenges}
Developing a robust multi-hop QA system presents several interconnected challenges. One of the foremost difficulties lies in accurately identifying and selecting the two relevant passages from ten candidate documents\cite{HotPotQA}. This requires leveraging both dense semantic embeddings and lexical signals while minimizing the inclusion of irrelevant or redundant content. Equally important is ensuring that the selected documents collectively offer distinct yet complementary pieces of evidence, an essential requirement for effective multi-hop reasoning. Another critical obstacle involves the decomposition of complex multi-hop questions into simpler, more targeted sub-queries\cite{fudecomposing}. This process, referred to as query optimization pipeline, must preserve the contextual integrity of the original query to maintain answer fidelity. In addition, the efficient computation of embeddings poses a computational challenge, particularly when striving to balance representational richness with runtime feasibility. To address this, fallback strategies such as integrating high-capacity models like OllamaEmbeddings with lighter alternatives like MiniLM are essential. Finally, generating coherent and precise answers with the correct format from multiple retrieved contexts, especially when the available information is partial or ambiguous, requires advanced prompt engineering techniques to regulate the output of the model.

\subsection{Contributions}
Our work addresses these challenges through a series of focused and impactful contributions. We begin by implementing and systematically comparing three retrieval strategies, the cosine similarity, Maximal Marginal Relevance (MMR), and a novel hybrid approach, within a unified and extensible retrieval framework. The hybrid method we propose combines dense vector embeddings with keyword-based lexical matching, enhanced by an MMR-based re-ranking mechanism that balances relevance and diversity. This approach leads to improved performance in multi-hop QA settings, where comprehensive and non-redundant evidence is essential. To further strengthen retrieval precision, we introduce refined query optimization pipeline techniques that effectively break down complex multi-hop queries into more tractable sub-queries without compromising contextual integrity. Additionally, we develop a robust embedding system that prioritizes OllamaEmbeddings for high-quality contextual representation, while integrating a fallback to the more computationally lightweight MiniLM to ensure system adaptability and resilience across environments. Finally, our contributions are validated through extensive experimentation on the HotpotQA dataset, where we evaluate retrieval performance using Exact Match and F1 scores, and perform detailed error analysis to identify strengths, limitations, and avenues for future enhancement.

Kunlong Zhang was responsible for the implementation of data embedding strategies and baseline models used in the system. Zichen Zhang developed the hybrid retrieval approach, integrating dense and lexical signals with MMR-based ranking. Hongwei Ruan led the implementation of the query optimization pipeline component, adapting and extending the EfficientRAG framework for query decomposition. Yiming Luo designed the experimental setup and conducted the result visualization and performance analysis.

\section{Related Work}

\subsection{Traditional Question Answering Systems}

Early question answering (QA) systems primarily relied on keyword matching and rule-based approaches, which often struggled with contextual understanding and complex language structures. With limited capacity for semantic interpretation, these systems were typically constrained to single-hop information retrieval. The advent of large-scale annotated datasets, such as HotpotQA, significantly advanced the field by enabling the development of data-driven models that learn intricate answer extraction patterns from extensive corpora.

\subsection{Transformer-Based Models for QA}

The introduction of Transformer-based architectures has revolutionized machine comprehension by enabling models to capture complex linguistic patterns and contextual relationships. Qwen 2.5\cite{qwen}, a state-of-the-art Transformer-based language model developed by Alibaba, exemplifies this leap in performance. Building upon the foundations of earlier models, Qwen 2.5 incorporates architectural optimizations, enhanced training strategies, and extensive multilingual pre-training to achieve superior generalization and efficiency. Its variants, such as Qwen-1.8B, Qwen-7B, offer scalable options suited for different computational budgets and deployment scenarios. Despite their impressive performance, Transformer-based models still face challenges in multi-hop reasoning, which requires integrating multiple pieces of evidence to formulate coherent answers. In our study, we evaluate Qwen-2.5-7B among other architectures to benchmark their effectiveness in extractive QA tasks.

\subsection{Retrieval-Augmented Generation for Multi-Hop QA}

Recent advancements have explored retrieval-augmented approaches to overcome the limitations of parametric models in addressing complex, multi-hop QA tasks. Retrieval-Augmented Generation (RAG), initially introduced by Facebook AI\cite{Facebook}, integrates a retrieval module with a generative model, such as BART\cite{BART} or Qwen, to dynamically access and incorporate information from vast document repositories. In the context of multi-hop QA, where answering questions requires sequential reasoning over interconnected evidence, RAG-based systems offer a promising solution. By leveraging both retrieval and generation, these systems can retrieve diverse information fragments and synthesize them into coherent, contextually rich responses. Our work focuses on enhancing RAG specifically for multi-hop scenarios, aiming to improve the chaining of evidence and overall answer accuracy.

\subsection{EfficientRAG}

A recent advancement in multi-hop retrieval is EfficientRAG, introduced by Zhuang et al. \cite{zhuang2024efficientragefficientretrievermultihop}. Unlike traditional RAG methods that rely on multiple large language model (LLM) invocations for iterative query generation, EfficientRAG proposes a lightweight and efficient retrieval framework. It utilizes a dual-component architecture, a Labeler \& Tagger and a Filter, which together iteratively refine and reformulate queries without invoking LLMs at each step. The Labeler identifies salient tokens in retrieved chunks that are potentially useful for answering the question, while the Filter constructs new queries using these tokens to facilitate further retrieval rounds. This iterative approach significantly reduces computational cost and inference latency while maintaining or exceeding performance benchmarks across datasets like HotpotQA, MuSiQue, and 2WikiMQA. EfficientRAG demonstrates strong recall with fewer retrieved chunks and outperforms many LLM-based iterative retrievers in both end-to-end accuracy and cross-domain transferability. Its plug-and-play design also offers practical benefits for scalable and cost-efficient multi-hop QA systems.

\section{Dataset}
We use \textbf{HotpotQA} to evaluate the fundamental capabilities of transformer-based QA models. HotpotQA is a challenging and widely adopted question answering dataset designed to promote multi-hop reasoning. Unlike traditional datasets such as SQuAD, where answers are typically derived from a single passage, HotpotQA requires models to integrate information across multiple supporting documents to derive an answer. This dataset includes both factoid questions and supporting sentence annotations, encouraging deeper comprehension and reasoning capabilities. Its emphasis on multi-hop inference and evidence-based reasoning makes HotpotQA particularly suitable for evaluating a model's ability to perform complex information synthesis, thus offering a more rigorous benchmark for QA system evaluation.

The dataset contains around 113k question-answer pairs. The key features of HotpotQA include:
\begin{itemize}
    \setlength{\itemsep}{-4.5pt}
    \item \textbf{Multi-hop Reasoning}: Models must piece together evidence from different contexts.
    \item \textbf{Supporting Facts}: HotpotQA provides annotations of the sentences necessary for answering each question.
    \item \textbf{Rigorous Evaluation}: Its complexity, including distractor passages and diverse question types, challenges models to accurately distinguish relevant information from noise.
\end{itemize}
    \includegraphics[width=\linewidth]{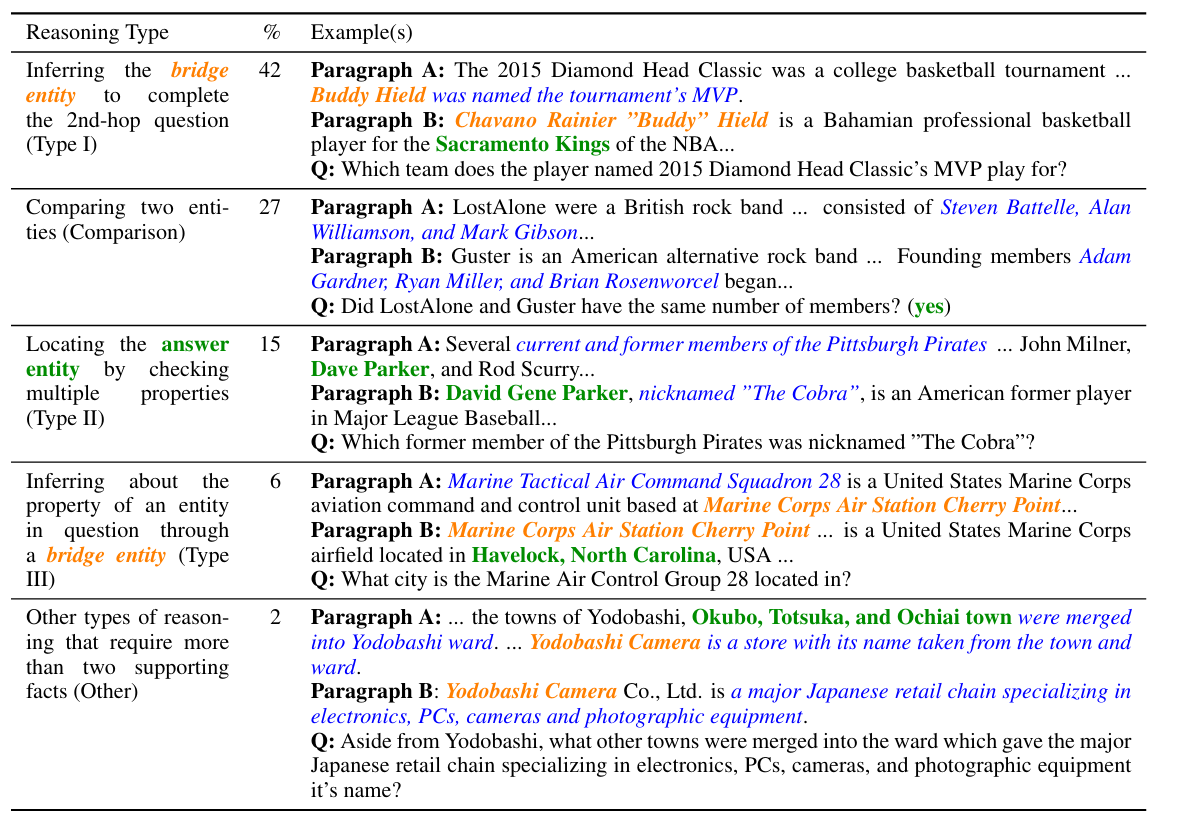}

\section{Methodology}

\subsection{Embedding Model Architecture}

Our system utilizes a dual-source embedding strategy:

\begin{itemize}
    \item \textbf{Primary}: Ollama's implementation of Qwen2.5 (7B parameter model) for generating high-quality contextual embeddings
    \item \textbf{Fallback}: SentenceTransformers' all-MiniLM-L6-v2 model, automatically activated if the primary model is unavailable
\end{itemize}

This fallback mechanism ensures system robustness and consistent performance across different computing environments. The embedding dimension varies based on the active model (1536 for Qwen2.5, 384 for MiniLM).

\subsection{Retrieval-Augmented Generation Framework}

Our question answering system for HotpotQA is built on a Retrieval-Augmented Generation (RAG) framework. The pipeline consists of three main components:
\begin{enumerate}
    \item Document retrieval from context pools
    \item Comparative analysis of multiple retrieval strategies
    \item Answer generation using retrieved context
\end{enumerate}

\subsection{Query Optimization Pipeline}
To further improve multi-hop retrieval effectiveness, we adopt and modify the EfficientRAG framework proposed by Zhuang. EfficientRAG introduces a lightweight, efficient retrieval system that replaces repeated large language model (LLM) calls with two specialized components: a Labeler and a Filter. The Labeler learns to identify salient tokens within retrieved passages, which are then used by the Filter to construct and refine follow-up queries. This architecture supports iterative multi-hop retrieval without invoking LLMs at every hop, reducing computational overhead while maintaining retrieval quality. In our implementation, we retain the core structure of EfficientRAG but make several modifications to better suit our system. These include customizing the token labeling thresholds, refining the query construction logic in the Filter module, and adapting the pipeline for compatibility with the HotpotQA dataset and our hybrid retrieval framework. We also integrate this modified EfficientRAG module with our embedding fallback mechanism to ensure robustness across varying query complexities. This integration contributes to more contextually relevant and computationally efficient retrieval, ultimately leading to enhanced end-to-end QA performance.

\subsection{Document Retrieval Approaches}

We implemented and compared three distinct retrieval methods to evaluate their effectiveness on multi-hop reasoning questions:

\subsubsection{Cosine Similarity Retrieval}
The most straightforward approach utilizes cosine similarity between query and document embeddings to identify relevant documents:

\begin{equation}
    \text{similarity}(q, d) = \frac{q \cdot d}{||q|| \cdot ||d||}
\end{equation}

Where $q$ is the query embedding and $d$ is each document embedding. We select the top-$k$ documents with highest similarity scores. This method serves as our baseline and excels at identifying documents with high semantic relevance to the query.

\subsubsection{Maximal Marginal Relevance (MMR)}
To address the problem of redundancy in retrieved documents, we implemented MMR which balances relevance with diversity:

\begin{equation}
    \text{MMR} = \arg\max_{d_i \in R \setminus S} \left[ \lambda \cdot \text{sim}(q, d_i) - (1 - \lambda) \cdot \max_{d_j \in S} \text{sim}(d_i, d_j) \right]
\end{equation}

Where:
\begin{itemize}
    \item $R$ is the set of candidate documents
    \item $S$ is the set of already selected documents
    \item $\lambda$ controls the trade-off between relevance and diversity
\end{itemize}

With $\lambda = 0.5$, MMR selects documents that are not only relevant to the query but also contain unique information compared to previously selected documents. This approach is particularly valuable for multi-hop reasoning where diverse information sources are crucial.

\subsubsection{Hybrid Retrieval}
We also implemented a hybrid approach proposed by \cite{sawarkar2024blended} that combines dense embedding similarity with lexical matching to capture both semantic and keyword-based relevance:

\begin{enumerate}
    \item Calculate embedding similarity between query and documents.
    \item Calculate lexical overlap score based on matching terms between query and document.
    \item Combine scores: $\text{combined\_score} = \text{embedding\_similarity} + \text{lexical\_score}$.
    \item Apply MMR using the combined scores to select diverse yet relevant documents.
    \item Re-rank the selected documents using the MMR-based scores to further refine the retrieval order.
\end{enumerate}

This hybrid method addresses limitations of pure embedding-based approaches by incorporating explicit term matching, which is particularly effective for named entities and rare terms common in HotpotQA. The additional re-ranking step ensures that the final set of retrieved documents maintains both high relevance and diversity. 

All experiments use top-$k=2$ to match HotpotQA's structure of having two supporting documents with 8 noise documents per question.

\subsection{Answer Generation}

For answer generation, we employ a constrained generation approach using the Ollama API with carefully crafted prompts:

\begin{enumerate}
    \item Retrieved documents are concatenated to form the context
    \item System prompts instruct the model to:
    \begin{itemize}
        \item Produce only ``yes'', ``no'', or a specific text span for factual information
        \item Return ``no answer'' when insufficient information is available
        \item Maintain lowercase formatting for consistency
    \end{itemize}
\end{enumerate}

We utilize a low temperature setting (0.1) to encourage deterministic, factual responses rather than creative generation.

\section{Experiments}

\subsection{Experimental Setup}

We evaluate our system using the HotpotQA development set. All retrieval methods, Cosine Similarity, Maximal Marginal Relevance (MMR), and our proposed Hybrid approach, are evaluated under consistent parameters. The test set consists of 1000 questions, ensuring a balanced and interpretable assessment. Our Hybrid retrieval pipeline is integrated with query decomposition, token labeling, next-hop query filtering, negative sampling, and synthesized training data.

We report performance using two standard metrics: \textbf{Exact Match (EM)} and \textbf{F1} Score. EM measures the proportion of answers that exactly match the gold label, while F1 captures the harmonic mean of precision and recall, allowing for partial correctness. These metrics are widely adopted in QA evaluations such as SQuAD and HotpotQA.

\begin{figure}[h!]
\includegraphics[width=\linewidth]{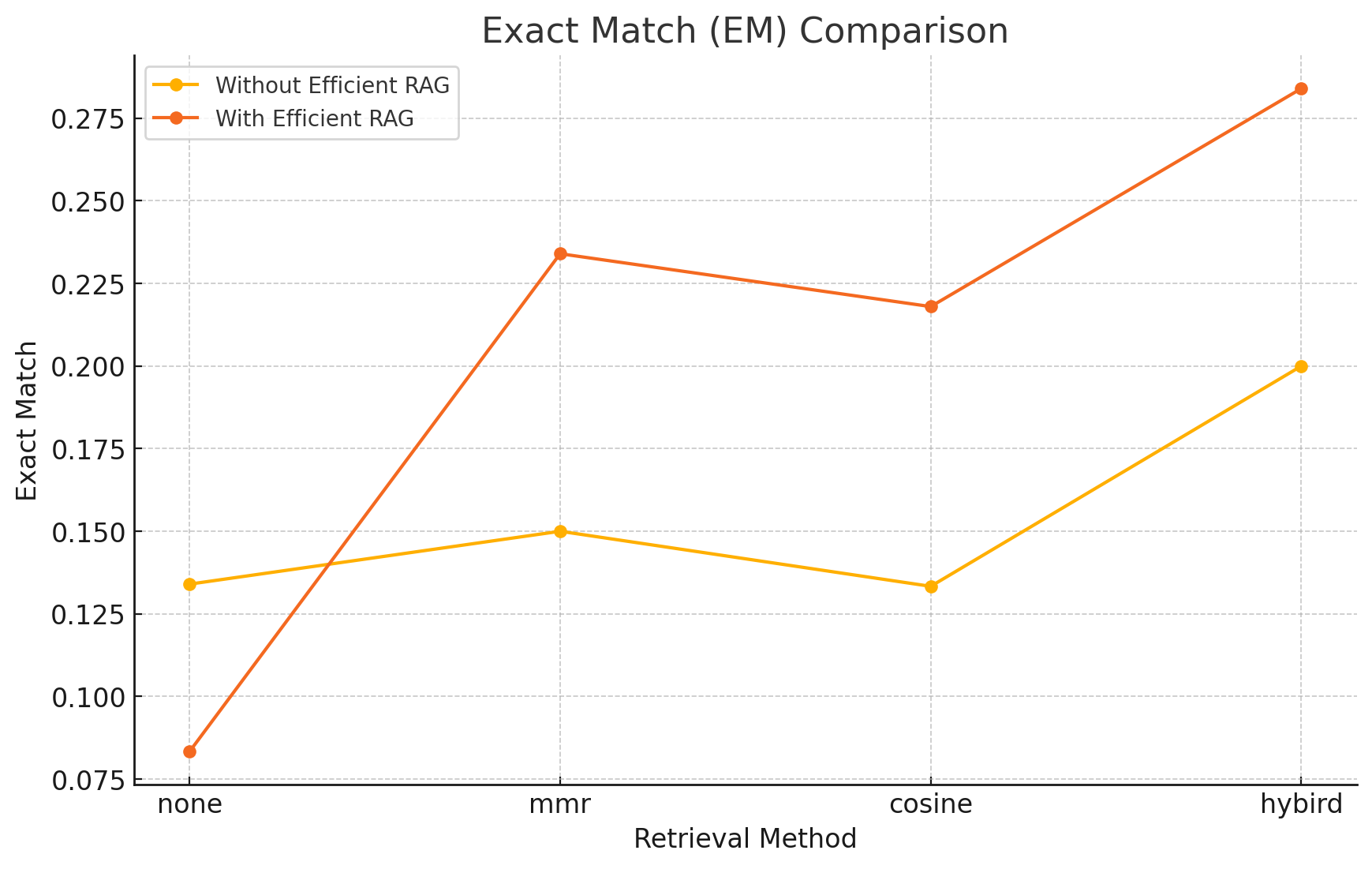}
\caption{Comparison of Exact Match Across Retrieval Methods}
\end{figure}

\begin{figure}[h!]
\includegraphics[width=\linewidth]{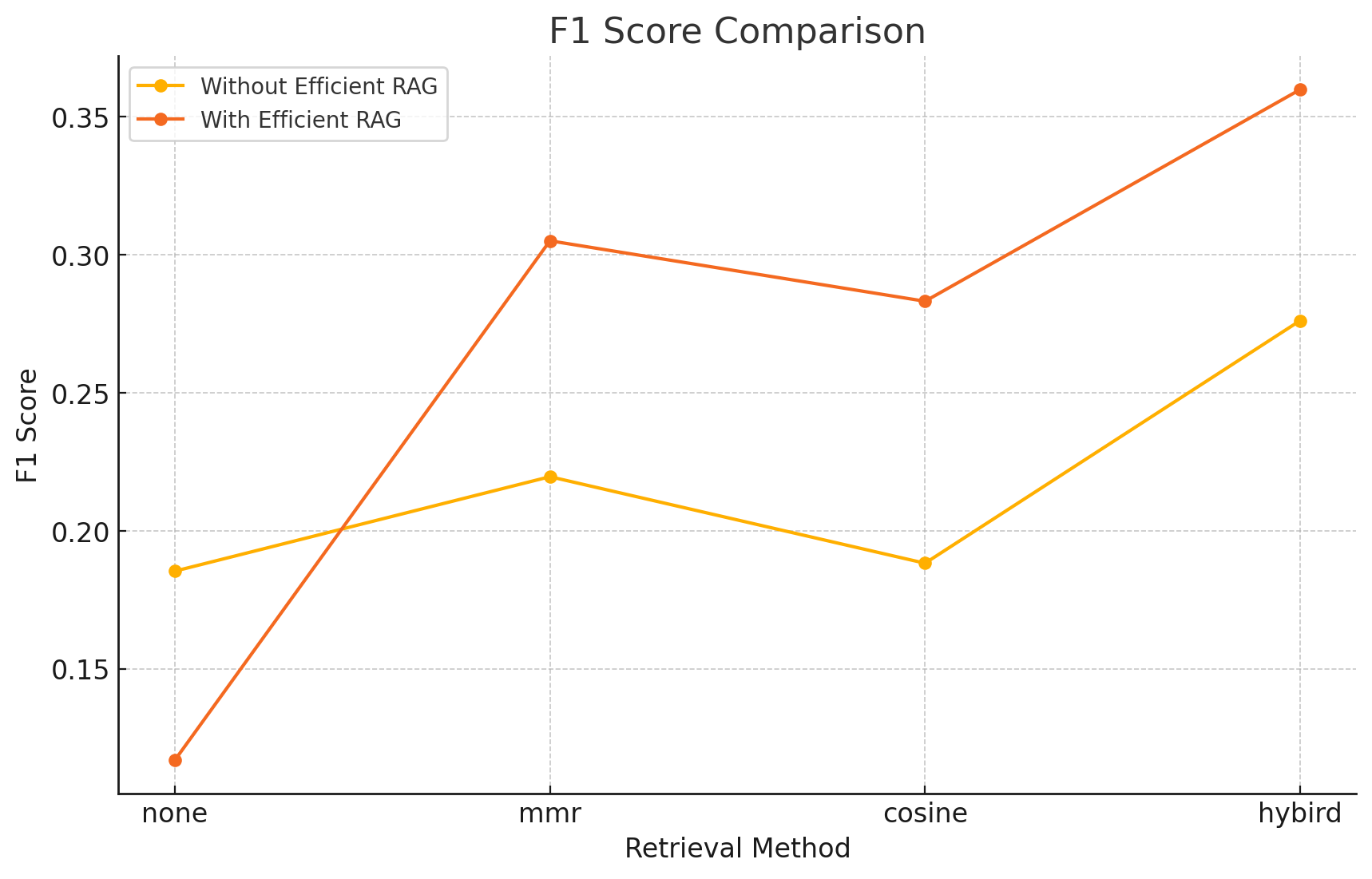}
\caption{Comparison of F1 Scores Across Retrieval Methods}
\end{figure}

\subsection{Retrieval Method Performance}

The Hybrid approach demonstrates the strongest performance, achieving an EM of 0.200 and an F1 of 0.276. This represents a 50\% relative improvement in EM compared to Cosine Similarity (0.200 vs. 0.133) and a 33\% improvement over MMR (0.200 vs. 0.150). Similarly, F1 scores improve by 46.8\% over Cosine (0.276 vs. 0.188) and 25.5\% over MMR (0.276 vs. 0.220), indicating better coverage of the ground truth answer.

MMR outperforms Cosine Similarity across all metrics, highlighting the benefits of diversity-aware document selection in multi-hop QA. These results confirm that enriching semantic retrieval with lexical overlap and controlled redundancy leads to more accurate and interpretable answers.

\section{Evaluation}

\subsection{Quantitative Insights}
In our quantitative evaluation, the Hybrid retrieval method demonstrates substantial gains over both cosine‐similarity and MMR baselines.  Specifically, when compared to a pure cosine‐similarity retriever, Hybrid achieves a 50\% relative increase in Exact Match (EM) and a 47\% relative increase in F1 score.  Against the MMR‐only approach, Hybrid still secures a 33\% relative improvement in EM and a 25.5\% uplift in F1.  Notably, MMR by itself yields a 13\% gain in EM and a 17\% gain in F1 over the cosine baseline, highlighting the diversity‐driven benefits of maximal marginal relevance pruning.  These results confirm that the fusion of dense embeddings, lexical overlap scoring, and MMR ranking produces superior quantitative performance.

\subsection{Qualitative Analysis}

\paragraph{Strengths of the Hybrid Approach}
The Hybrid retriever integrates dense embedding similarity with explicit keyword overlap signals and MMR pruning in a single pipeline.  This integration enhances entity recall by ensuring that passages containing rare or domain‐specific terms are retrieved even when embeddings alone would miss them.  By combining embedding and lexical signals, it also promotes evidence complementarity: each selected passage contributes unique facts necessary for constructing complete multi‐hop reasoning chains, thereby minimizing redundancy.  Finally, the MMR stage systematically removes low‐value or duplicate passages, increasing precision while maintaining high recall.  These strengths are particularly evident on queries requiring comparative judgments or multi‐step inference.

\paragraph{Advantages and Limitations of MMR}
The MMR‐only retriever enforces diversity constraints that broaden contextual coverage and improve recall on multi‐hop questions by aggregating evidence from dispersed sources.  However, without explicit lexical matching, MMR may underperform on domain‐specific or low‐frequency queries, as it can overlook critical keywords that appear sparsely in the corpus.

\subsection{Error Analysis}
Despite its strong performance, the Hybrid retrieval method exhibits several limitations. First, distractor bias can lead to retrieval of passages that share high lexical overlap but contain irrelevant content, thereby misleading the answer generation module with spurious “supporting” evidence, this issue is exacerbated when rare terms coincide across unrelated topics. Second, purely retrieval‑based pipelines lack specialized mechanisms for complex reasoning tasks, such as temporal sequencing or nuanced comparative judgments, meaning that although the correct facts may be retrieved, the system cannot reliably assemble them into a coherent, correctly ordered answer. Third, the pipeline is sensitive to token‑labeling thresholds and query‑decomposition heuristics; these parameters can cause critical tokens to be omitted or noisy tokens to be introduced, resulting in suboptimal follow‑up queries, and fine‑tuning them for different domains remains a challenging task that can affect retrieval consistency across question types.

\subsection{Comparison to State‑of‑the‑Art}
While supervised, fine‑tuned architectures on HotpotQA (for example, HGN reporting sp\_f1=0.789) achieve higher end‑to‑end accuracy, our zero‑shot pipeline offers three key practical advantages.  First, it is immediately applicable without any task‑specific fine‑tuning, enabling rapid deployment across new domains.  Second, the clear retrieval chains provide explainability, allowing practitioners to trace and interpret each supporting passage.  Third, the inference cost is substantially lower than end‑to‑end LLM fine‑tuning, making the approach well suited for low‑resource and domain‑adaptive question‐answering scenarios.  These characteristics render the Hybrid retriever a versatile and efficient solution for zero‑shot multi‑hop question answering.

\section{Conclusion}

In this work, we presented a comprehensive evaluation of multi‑hop question answering models by benchmarking cosine similarity and Maximal Marginal Relevance against our novel Hybrid retrieval strategy, which is further enhanced by a query optimization pipeline incorporating query decomposition, token labeling, next‑hop filtering, and negative sampling. Experiments on HotpotQA demonstrate that the Hybrid approach achieves substantial performance gains over baseline methods (EM=0.200, F1=0.276) while preserving explainability and maintaining low computational overhead. Detailed error analysis revealed distractor bias, limited retrieval depth, and challenges in complex temporal reasoning, motivating future work on deeper relevance modeling and the integration of specialized temporal and comparative reasoning modules. The plug‑and‑play design of our pipeline ensures seamless adaptation to new domains and datasets without task‑specific fine‑tuning. Overall, our Hybrid RAG framework strikes an effective balance between efficiency, interpretability, and accuracy, offering a practical zero‑shot solution for multi‑hop QA and laying the foundation for further advances in retrieval‑augmented systems.

\section{GitHub repository} 

\href{https://github.com/Rhongomyniadz/rag_evaluation}{\texttt{github.com/Rhongomyniadz/rag\_evaluation}}

\nocite{zhuang2024efficientragefficientretrievermultihop}

\bibliography{example_paper}

\begin{thebibliography}{10}
\providecommand{\natexlab}[1]{#1}
\providecommand{\url}[1]{\texttt{#1}}
\expandafter\ifx\csname urlstyle\endcsname\relax
  \providecommand{\doi}[1]{doi: #1}\else
  \providecommand{\doi}{doi: \begingroup \urlstyle{rm}\Url}\fi

\bibitem[Devlin et~al.(2019)Devlin, Chang, Lee, and Toutanova]{BERT}
Devlin, J., Chang, M.-W., Lee, K., and Toutanova, K.
\newblock Bert: Pre-training of deep bidirectional transformers for language understanding, 2019.
\newblock URL \url{https://arxiv.org/abs/1810.04805}.

\bibitem[Fu et~al.(2021)Fu, Wang, Zhang, Zhou, and Yan]{fudecomposing}
Fu, R., Wang, H., Zhang, X., Zhou, J., and Yan, Y.
\newblock Decomposing complex questions makes multi-hop qa easier and more interpretable, 2021.
\newblock URL \url{https://arxiv.org/abs/2110.13472}.

\bibitem[Lewis et~al.(2019)Lewis, Liu, Goyal, Ghazvininejad, Mohamed, Levy, Stoyanov, and Zettlemoyer]{BART}
Lewis, M., Liu, Y., Goyal, N., Ghazvininejad, M., Mohamed, A., Levy, O., Stoyanov, V., and Zettlemoyer, L.
\newblock Bart: Denoising sequence-to-sequence pre-training for natural language generation, translation, and comprehension, 2019.
\newblock URL \url{https://arxiv.org/abs/1910.13461}.

\bibitem[Lewis et~al.(2021)Lewis, Perez, Piktus, Petroni, Karpukhin, Goyal, Küttler, Lewis, tau Yih, Rocktäschel, Riedel, and Kiela]{Facebook}
Lewis, P., Perez, E., Piktus, A., Petroni, F., Karpukhin, V., Goyal, N., Küttler, H., Lewis, M., tau Yih, W., Rocktäschel, T., Riedel, S., and Kiela, D.
\newblock Retrieval-augmented generation for knowledge-intensive nlp tasks, 2021.
\newblock URL \url{https://arxiv.org/abs/2005.11401}.

\bibitem[Qwen et~al.(2025)Qwen, :, Yang, Yang, Zhang, Hui, Zheng, Yu, Li, Liu, Huang, Wei, Lin, Yang, Tu, Zhang, Yang, Yang, Zhou, Lin, Dang, Lu, Bao, Yang, Yu, Li, Xue, Zhang, Zhu, Men, Lin, Li, Tang, Xia, Ren, Ren, Fan, Su, Zhang, Wan, Liu, Cui, Zhang, and Qiu]{qwen}
Qwen, :, Yang, A., Yang, B., Zhang, B., Hui, B., Zheng, B., Yu, B., Li, C., Liu, D., Huang, F., Wei, H., Lin, H., Yang, J., Tu, J., Zhang, J., Yang, J., Yang, J., Zhou, J., Lin, J., Dang, K., Lu, K., Bao, K., Yang, K., Yu, L., Li, M., Xue, M., Zhang, P., Zhu, Q., Men, R., Lin, R., Li, T., Tang, T., Xia, T., Ren, X., Ren, X., Fan, Y., Su, Y., Zhang, Y., Wan, Y., Liu, Y., Cui, Z., Zhang, Z., and Qiu, Z.
\newblock Qwen2.5 technical report, 2025.
\newblock URL \url{https://arxiv.org/abs/2412.15115}.

\bibitem[Rajpurkar et~al.(2016)Rajpurkar, Zhang, Lopyrev, and Liang]{SQuADQA}
Rajpurkar, P., Zhang, J., Lopyrev, K., and Liang, P.
\newblock Squad: 100,000+ questions for machine comprehension of text, 2016.
\newblock URL \url{https://arxiv.org/abs/1606.05250}.

\bibitem[Sawarkar et~al.(2024)Sawarkar, Mangal, and Solanki]{sawarkar2024blended}
Sawarkar, K., Mangal, A., and Solanki, S.~R.
\newblock Blended rag: Improving rag (retriever-augmented generation) accuracy with semantic search and hybrid query-based retrievers.
\newblock In \emph{2024 IEEE 7th International Conference on Multimedia Information Processing and Retrieval (MIPR)}, volume~24, pp.\  155--161. IEEE, 2024.
\newblock \doi{10.1109/mipr62202.2024.00031}.

\bibitem[Xia et~al.(2015)Xia, Xu, Lan, Guo, and Cheng]{MMR}
Xia, L., Xu, J., Lan, Y., Guo, J., and Cheng, X.
\newblock Learning maximal marginal relevance model via directly optimizing diversity evaluation measures.
\newblock In \emph{Proceedings of the 38th International ACM SIGIR Conference on Research and Development in Information Retrieval}, SIGIR '15, pp.\  113–122, New York, NY, USA, 2015. Association for Computing Machinery.
\newblock ISBN 9781450336215.
\newblock \doi{10.1145/2766462.2767710}.
\newblock URL \url{https://doi.org/10.1145/2766462.2767710}.

\bibitem[Yang et~al.(2018)Yang, Qi, Zhang, Bengio, Cohen, Salakhutdinov, and Manning]{HotPotQA}
Yang, Z., Qi, P., Zhang, S., Bengio, Y., Cohen, W.~W., Salakhutdinov, R., and Manning, C.~D.
\newblock Hotpotqa: A dataset for diverse, explainable multi-hop question answering, 2018.
\newblock URL \url{https://arxiv.org/abs/1809.09600}.

\bibitem[Zhuang et~al.(2024)Zhuang, Zhang, Cheng, Yang, Liu, Huang, Lin, Rajmohan, Zhang, and Zhang]{zhuang2024efficientragefficientretrievermultihop}
Zhuang, Z., Zhang, Z., Cheng, S., Yang, F., Liu, J., Huang, S., Lin, Q., Rajmohan, S., Zhang, D., and Zhang, Q.
\newblock Efficientrag: Efficient retriever for multi-hop question answering, 2024.
\newblock URL \url{https://arxiv.org/abs/2408.04259}.

\end{thebibliography}
\bibliographystyle{icml2025}

%%%%%%%%%%%%%%%%%%%%%%%%%%%%%%%%%%%%%%%%%%%%%%%%%%%%%%%%%%%%%%%%%%%%%%%%%%%%%%%
%%%%%%%%%%%%%%%%%%%%%%%%%%%%%%%%%%%%%%%%%%%%%%%%%%%%%%%%%%%%%%%%%%%%%%%%%%%%%%%
% APPENDIX
%%%%%%%%%%%%%%%%%%%%%%%%%%%%%%%%%%%%%%%%%%%%%%%%%%%%%%%%%%%%%%%%%%%%%%%%%%%%%%%
%%%%%%%%%%%%%%%%%%%%%%%%%%%%%%%%%%%%%%%%%%%%%%%%%%%%%%%%%%%%%%%%%%%%%%%%%%%%%%%
\newpage
\appendix
\onecolumn
%%%%%%%%%%%%%%%%%%%%%%%%%%%%%%%%%%%%%%%%%%%%%%%%%%%%%%%%%%%%%%%%%%%%%%%%%%%%%%%
%%%%%%%%%%%%%%%%%%%%%%%%%%%%%%%%%%%%%%%%%%%%%%%%%%%%%%%%%%%%%%%%%%%%%%%%%%%%%%%

\end{document}